# Clustering and Classification with Non-Existence Attributes: A Sentenced Discrepancy Measure Based Technique

Y. A. Joarder[1], Emran Hossain[2], Al Faisal Mahmud[3]

[1,2,3]Department of Computer Science & Engineering (CSE)

World University of Bangladesh (WUB), Dhaka, Bangladesh

**Abstract.** For some or all of the data instances a number of independent-world clustering issues suffer from incomplete data characterization due to losing or absent attributes. Typical clustering approaches cannot be applied directly to such data unless pre-processing by techniques like imputation or marginalization. We have overcome this drawback by utilizing a Sentenced Discrepancy Measure which we refer to as the Attribute Weighted Penalty based Discrepancy (AWPD). Using the AWPD measure, we modified the K-MEANS++ and Scalable K-MEANS++ for clustering algorithm and k Nearest Neighbor (kNN) for classification so as to make them directly applicable to datasets with non-existence attributes. We have presented a detailed theoretical analysis which shows that the new AWPD based K-MEANS++, Scalable K-MEANS++ and kNN algorithm merge into a local prime among the number of iterations is finite. We have reported in depth experiments on numerous benchmark datasets for various forms of Non-Existence showing that the projected clustering and classification techniques usually show better results in comparison to some of the renowned imputation methods that are generally used to process such insufficient data. This technique is designed to trace invaluable data to: directly apply our method on the datasets which have Non-Existence attributes and establish a method for detecting unstructured Non-Existence attributes with the best accuracy rate and minimum cost.

## 1    Introduction

In data analytics, clustering is a fundamental technique which helps to partition a given dataset into healthier groups as well as makes some groups among the data instances with the relative similarity. In general, clustering is used in unsupervised learning as working with nonclass label data. Clustering algorithms attempt to partition a collection of data instances (characterized by some attributes), into completely different clusters specifying the member instances of any given clusters complement one another and they are different from the members of the opposite cluster. In clustering, by the use of suitable algorithm, the similar and dissimilar data create their own groups [1].

On the other side, classification is a fundamental technique which helps to classify the unobserved data in a given dataset into some classified groups based on relative similarity among the data instances. In addition, classification is generally used in supervised learning as working with class label data.

Clustering and Classification techniques both are extensively used and hence being constantly investigated in statistics, machine learning, and pattern recognition. Clustering and Classification algorithms find applications in the different sector, for example: banking, space research, economics, electronic design, and marketing. It creates a problem of clustering or classification when the dataset presents with the Non-Existence attribute. If we try to cluster or classify in the datasets with a Non-Existence attribute, we will get some inefficient issues such as creating some empty sets and get extra Non-Existence at the end of clustering or classifying in the given dataset [1]. For example, clustering and classification has been using for grouping connected documents in web browsing [2], classification has been used to trace suspicious (possibly fraudulent) behavior on the basis of previous transactions of customers in banking system, [3], for formulating effective marketing strategies, it is possible to group or cluster the same types of customers according to their choice of products by using clustering [4], both clustering and classification techniques have been used for distinguishing dangerous zones on the basis of previous geographical point locations in earthquake [5], [6], [7]. However, once we analyze such real-world datasets, we could tend to encounter incomplete data wherever some attributes of a number of the data instances are Non-Existence . For example, web documents could have some invalid hyperlinks. Such Non-Existence is also vital because of a range of reasons like data input errors, inaccurate mensuration, instrumentality malfunction or limitations, and mensuration noise or data corruption and so on. These categories are define as unstructured Non-Existence [8], [9]. Instead, on the contrary, in structural Non-Existence , all the attributes are not made public for all the data instances inside the dataset. These categories are termed as structural

Non-Existence or absence of attributes [10]. For example of structural Non-Existence , credit-card details might not be outlined for non-credit card users of a bank.

The Non-Existence Attribute is an unaccepted data into a dataset where some data instances are Non-Existence . Non-Existences attributes are also called Non-Existence attributes. For researchers, maintaining Non-Existence Attributes has always been a challenge because common learning approaches cannot be directly applied to such inaccurate data, without appropriate preprocessing: Imputation and Marginalization. Once the rate of Non-Existence is low, the data instances with Non-Existence values could also be unnoticed. This approach is called marginalization. Marginalization cannot be applied to data having a large range of Non-Existence values, because it might result in the loss of a large quantity of data. Therefore, sophisticated methods are needed to fill in the vacancies within the data, in order that ancient learning methods may be applied afterward. However, inferences drawn from data having an oversized fraction of Non-Existence values could also be severely crooked, despite the utilization of such sophisticated imputation methods [11].

## 1.1 Contributions of this research are:

- Formulating K-MEANS++ and Scalable K-MEANS++ clustering and kNN classification problem for datasets with non-existence attributes focused on proposed AWPD.
- Providing a new approach named Sentenced Discrepancy Measure (SDM).
- Proposing the use of SDM called the Attribute Weighted Penalty based Discrepancy.
- Developing those algorithms and approach to prove the new formulation.
- Proving the propose algorithm is modified K-MEANS++, Scalable K-MEANS++ and kNN optimization problem formulated with the AWPD measure.
- Providing a details of all algorithms for simulating four types of Non-Existence, namely MCAR, MAR, MNAR-1 and MNAR-2.
- Showing the results through tables, bar charts and line graph.

## 2 . Literature Review

To decide how to handle non-existence data, one needs to know why data are Non-Existence . There are three types of Non-Existence mechanism [9]. These are: Non-Existence Completely at Random (MCAR), Non-Existence at Random (MAR), Non-Existence Not at Random (MNAR). In case of MCAR, Non-Existence value is not depending on both observed and unobserved data. For example of MCAR, A citizen is unable to participate due to reasons unrelated to the survey, like traffic or schedule issues. In case of MAR, Non-Existence attributes are depending on observed attributes but cannot depend on unobserved attributes. For example of MAR, College-goers are less doubtless to report their financial gain than office-goers. However, whether or not a college-goer can report his/her financial gain is freelance of the particular financial gain. MNAR refers to the case wherever Non-Existence is subject to the unobserved attributes of association in nursing instance. As an example of MNAR, people with lower earnings are less doubtless to report their financial gains within the annual income survey.

After that [1] and [15] told that MNAR has two subtypes here is MNAR-1: this only builds on the unobserved attributes. Another one is MNAR-2: this term is builds on the both observed and unobserved attributes. Given datasets with Non-Existence attribute cannot be directly introduced in this dataset. Hence many researchers used imputation method and marginalization (pre-processing) [1]. These are: Zero imputation is the process of changing the dataset with Non-Existence values, it replaces by zero (0). Mean imputation is the process of replacing the dataset with Non-Existence values, first observed full dataset and distance measure by Euclidean distance then replacing the Non-Existence value. k Nearest Neighbor Imputation is applied wherever a Non-Existence attribute of a data instance is countable to have the type of resembling attributes of its k nearest neighbors (on the observed subspace) [12].

The k Nearest Neighbor (kNN) classifier is the distant and simplest pattern classification techniques. The kNN classifier does not build any previous assumptions concerning the category distributions [13]. The 1NN classifier achieves a chance of error but double the Bayes chance of error once the scale of the training set tends to eternity [14]. The kNN classifier functions by searching the K Nearest Neighbors of a check purpose from among a set of training data instances with best-known class labels [14].

When the given datasets Non-Existence is at a low level then the usage process is called marginalization. Marginalization means ignoring the Non-Existence [1].

In [1] and [15], they proposed that their proposed technique can be directly applied in datasets with Non-Existence attribute. K-MEANS FWPD is a technique which can be directly applied in a dataset with Non-Existence attribute for clustering without any pre-processing such as imputation or marginalization. FWPD is aided in K-MEANS algorithm. kNN-FWPD is a technique used for classification with Non-Existence attributes or absence attributes. FWPD is aided in kNN classifiers.

## 3. Methodology

### 3.1: Sentenced Discrepancy Measure (SDM)

Equation for SDM,

$$\delta d\left(a_i, a_j\right) = \sqrt{\sum_{i=1, j=1} \left(a_{i,l} - a_{j,l}\right)^2 * 1/2}$$

Example:

As determined earlier, one potential thanks to adapt supervised also as unsupervised learning ways to issues with Non-Existence is to change the space or difference measure underlying the training technique. The idea is that the changed Discrepancy measure ought to use the common observed attributes to produce approximations of the distances between the data instances if they were to be absolutely observed. PDM is one of the way. These methods do not need marginalization or imputation but are likely to produce better performances than both of these two. For example, let $P_{full} = \{p_1 = (1,5), p_2 = (2,3), p_3 = (3,6)\}$ be a dataset comprising of three points in R2. Then, we collect some value $d_E$ $(p_1, p_2) = \sqrt{5}$ and $d_E$ $(p_1, p_3) = \sqrt{5}$ (where $d_E$ $(p_i, p_j)$ being the Euclidean distance among any two fully observed points $p_i$ and $p_j$ in $P_{full}$. Guess, This first level associate (1, 5) be unobserved, now this is the incomplete dataset $P_{full} = \{p'_1 = (*,5), p_2 = (2,3), p_3 = (3,6)\}$ ('*') means a Non-Existence value), on which training need to be accomplish. Please remember this, that is the exception of unstructured Non-Existence (Even though the unrecognized value is familiar to exists), as opposite to each other of the structural Non-Existence [10]. We are using ZI, MI and 1NNI severally, we have procure the following refilled in datasets

$$P_{ZI} = \{p^\wedge_1 = (0,5), p_2 = (2,3), p_3 = (3,6)$$

$$P_{MI} = \{p^\wedge_1 = (2.5,5), p_2 = (2,3), p_3 = (3,6)$$

$$P_{1NNI} = \{p^\wedge_1 = (3,5), p_2 = (2,3), p_3 = (3,6)$$

PDM's incorrect calculations are due to the fact that the distance in the specific observed subspace does not represent the distance in the unobserved subspace [1].

$$\delta PDM\left(a_i, a_j\right) = \sqrt{\sum_{i=1,j=1}\left(a_{i,l} - a_{j,l}\right)^2} + \frac{1}{2}$$

Therefore, the discrepancies $\delta PDM(a_1, a_2)$ and $\delta PDM(a_1, a_3)$ are

$$\delta PDM(a_1, a_2) = \sqrt{(5-3)^2} + \frac{1}{2} = 2.5$$

$$\delta PDM(a_1, a_2) = \sqrt{(5-6)^2} + \frac{1}{2} = 1.5$$

After the observed distance through two data instances is effectively a lower bound on the Euclidean range among both (if completely observed), applying an appropriate penalty to this lower bound may contribute to a reasonable approximation of the actual distance. This method, named the Sentenced Discrepancy Measure (SDM), could resolve the drawback of PDM. The penalty between $p_1$ and $p_i$ can be calculated by the combination of the amount of attributes this are unobserved for at least one of the two data instances as well as the overall number of attributes throughout the dataset. Then, the Discrepancy $\delta$SDM $(p_1, p_i)$ between the Inherent calculation of $p_1$ and another $p_i \in$ P is

$$\delta SDM\left(a_i, a_j\right) = \sqrt{\sum_{i=1,j=1}\left(a_{i,l} - a_{j,l}\right)^2 + \frac{1}{2}}$$

Therefore, the discrepancies $\delta SDM(a_1, a_2)$ and $\delta SDM(a_1, a_3)$ are

$$\delta SDM(a_1, a_2) = \sqrt{(5-3)^2 * \frac{1}{2}} = 2.12$$

$$\delta SDM(a_1, a_3) = \sqrt{(5-6)^2 * \frac{1}{2}} = 1.22$$

### 3.2: Attribute Weighted Penalty based Discrepancy

Let the A $\subset \mathbb{R}^m$ dataset, that is- $A$ data instances are every characterized by $\mathbb{R}$ values m attributes. Then let $A$ comprise of n instances $a_i (i \in \{1,2,\ldots\ldots,n\})$, some that have attributes of non-existence. Let $\gamma_{a_i}$, contribute the set of attributes observed for $a_i$ data point. Subsequently, all of the set of attributes P=$\cup_{i=1}^{n} \gamma_{a_i}$ and $|P| = m$. The set of attributes observed for all data instances in $A$ is described as $\gamma_{obs} = \cap_{i=1}^{n} \gamma_{a_i} |\gamma_{obs}|$ may or may not be non zero. $\gamma_{miss=} P \backslash \gamma_{obs}$ The set of unobserved attributes which with at least one in $A$ data point.

**Mark 1:** Let, the distance among any two instances of data $a_i, a_j \in$ A in a subspace specified by $\gamma$ referred to as the $d_\gamma(a_i, a_j)$. Then, the distance observed between these two points' distances in the observed subspace can then be described as

$$d_{\gamma_{a_i} \cap \gamma_{a_j}}(a_i, a_j) = \sqrt{\sum_{l \epsilon \gamma_{a_i} \cap \gamma_{a_j}} \left(a_{i,l} - a_{j,l}\right)^2}$$

Where $a_{i,l}$ donates the $l$-th attribute of the data instance $a_i$. for convenience purposes, $d_{\gamma_{a_i} \cap \gamma_{a_j}}(a_i, a_j)$ is generalized to $d(a_i, a_j)$ in the reminder of this article.

**Mark 2:** If both $a_i$ and $a_j$ were to be fully observed, the Euclidean distance $d_E(a_i, a_j)$ between $a_i$ and $a_j$ would define as

$$d_E(a_i, a_j) = \sqrt{\sum_{l \epsilon P}(a_{i,l} - a_{j,l})^2}$$

**Mark 3:** The AWPD between $a_i$ and $a_j$ is defined as

$$q(a_i, a_j) = \frac{\sum_{l \epsilon P \backslash (\gamma_{a_i} \cap \gamma_{a_j})} w}{\sum_{l' \epsilon P} wl'}$$

**Final Mark:** The AWPD between $a_i$ and $a_j$ is

$$\delta(a_i, a_j) = (1 - \beta) + \frac{d(a_i, a_j)}{d_{max}} \times \beta \times q(a_i, a_j)$$

While β$\epsilon$(0,1) is a matric [1], that defines the relative importance of the two terms and $d_{max}$ is the maximum distance observed in their corresponding typical observed subspaces between any two points in $A$.

### 3.3: K-MEANS++ with AWPD

This portion introduces, using AWPD measure a reformulation of the K-MEANS clustering for non-existence datasets. Lloyd first proposed the standerd heuristic algorithm for solving the K-MEANS problem in 1957 [17]. The K-MEANS algorithms expands to a regional optimum of the non-convex simulation problem presented by the K-MEANS problem when the Euclidean distance between data points is the Discrepancy used [18]. Main problem of K-MEANS algorithm is initialization problem (randomly). After that the K-MEANS algorithm problem solved by K-MEANS++ algorithm. Its recover by K-MEANS++ algorithm (Smart initialization) [16].

The proposed K-MEANS++ problem of non-existence attribute datasets using the proposed AWPD measure, referred to as the K-MEANS++-AWPD issue. Therefore, the problem with K-MEANS++-AWPD partitioning the dataset A into k clusters can be formulated as follows:

P: minimize f (U, Z) $= \sum_{i=1}^{n} \sum_{j=1}^{k} ((1 - \beta) + \frac{d(a_i, a_j)}{d_{max}} \times \beta \times q(a_i, a_j))$

### 3.4: The K-MEANS++-AWPD Algorithm

To find the solution to the issue P, that is a non-convex problem program, we presented a heuristic Lloyds algorithm and resolved by [16] based on the AWPD (known as the K-MEANS++-AWPD algorithm) as follows:

1. Starting with a random initial cluster set $u$ such that $\sum_{j=1}^{k} u_{i,j} = 1$, set $t = 1$ and define the maximum number of smart initialized iterations.
2. Calculate the observed attributes of the cluster $C_j^t (1, 2, ..., k)$, of every cluster centroid $Z_j^t$. For all the data instances in the cluster $C_j^t$ having observed value for $l$-th attribute of a centroid $Z_j^t$ Should be the average of the corresponding attribute values. If for the attribute in question none of the data instances in $C_j^t$ observed values, it is essential to maintain the value $Z_{j,l}^{t-1}$ of the previous iteration attribute. Therefore, the attribute values are calculated as follow:

$$Z_{j,l}^t = \begin{cases} \left( \sum_{a_i \in A_l} u_{i,j}^t \times A_{i,l} \right) \Big/ \left( \sum_{a_i \in A_l} u_{i,j}^t \right), \forall \, l \cup_{a_i C_j^t} \gamma_{a_i} \\ \\ Z_{j,l}^{t-1}, \qquad \forall l \in \gamma_{a_j^{t-1}} \backslash \cup_{a_i C_j^t} \gamma_{a_i} \end{cases}$$

Here $A_l$ signifies the set of every $a_i \in A$ has observed attribute $l$ values.

3.  Give a subject $a_i (i = 1,2,\dots n)$ to every data point to the cluster relating to its closest centroid (in AWPD terms).

$$u_{i,j}^{t+1} = \begin{cases} 1, if \; Z_j^t = \arg\min_{z \epsilon Z^t} \delta(a_{i,Z}), \\ \\ null. \, otherwise \end{cases}$$

Set $t = t + 1$. If $U^t = U^{t-1}$ or $t = Maximum \; Iteration$ then go to step 4 otherwise go to step 2

4. Calculate the final cluster set $Z^*$ as:

$$Z_{j,l}^* = d \times \frac{\sum_{a_i \in A_l} u_{i,j}^{t+1} \times A_{i,l}}{\sum_{a_i \in A_l} u_{i,j}^{t+1}} \; \forall l \in \bigcup_{a_i \in C_j^{t+1}} \gamma_{a_i}$$

### 3.5: Scalable K-MEANS++ with AWPD

This portion introduces, using AWPD measure a reformulation of the scalable K-MEANS++ clustering for non-existences datasets. Lloyd first proposed the standard heuristic algorithm for solving the K-MEANS problem in 1957 [17]. The K-MEANS algorithms expands to a regional optimum of the non-convex simulation problem presented by the K-MEANS problem when the Euclidean distance between data points is the Discrepancy used [18]. Main problem of K-MEANS algorithm is initialization problem (randomly). After that, the K-MEANS algorithm problem solved by K-MEANS++ algorithm. It is recovered by K-MEANS++ algorithm (Smart initialization) [16]. However, K-MEANS++ have some problem, that means time complexity to large and calculation is high complexity. Its recover by scalable K-MEANS++ algorithm [19].

The proposed scalable K-MEANS++ problem of non-existence attribute datasets using the proposed AWPD measure, referred to as the scalable K-MEANS++-AWPD issue. Therefore, the problem with scalable K-MEANS++-AWPD partitioning the dataset A into k clusters can be formulated as follows:

P: minimize f $(U, Z) = \sum_{i=1}^{n} \sum_{j=1}^{k} ((1 - \beta) + \frac{d(a_i,a_j)d*l}{d_{max}} \times \beta \times q(a_i, a_j))$

### 3.6: Scalable K-MEANS++-AWPD Algorithm

To find the solution to the issue P, that is a non-convex problem program, we presented a heuristic Lloyds algorithm and resolved by [16] based on the AWPD (known as the K-MEANS++-AWPD algorithm) as follows:

1. Starting with a random initial cluster set $u$ such that $\sum_{j=1}^{k} u_{i,j} = 1$, set $t = 1$ and define the maximum number of smart initialized iterations.
2. Calculate the observed attributes of the cluster $C_j^t (1,2, ..., k)$, of every cluster centroid $Z_j^t$. For all the data instances in the cluster $C_j^t$ having observed value for $l$-th attribute of a centroid $Z_j^t$ Should be the average of the corresponding attribute values. If for the attribute in question none of the data instances in $C_j^t$ observed values, it is essential to maintain the value $Z_{j,l}^{t-1}$ of the previous iteration attribute. Therefore, the attribute values are calculated as follow:

$$Z_{j,l}^t = \begin{cases} (\sum_{a_i \in A_l} u_{i,j}^t \times A_{i,l}) \Big/ (\sum_{a_i \in A_l} u_{i,j}^t), \forall\, l \cup_{a_i C_j^t} \gamma_{a_i} \\ Z_{j,l}^{t\,1}\,, \qquad \forall l \in \gamma_{a_j^{t-1}} \backslash \cup_{a_i C_j^t} \gamma_{a_i} \end{cases}$$

Here $A_l$ signifies the set of every $a_i \in A$ has observed attribute$l$ values.

3. Give a subject $a_i (i = 1,2, ... n)$ to every data point to the cluster relating to its closest centroid (in AWPD terms).

$$u_{i,j}^{t+1} = \begin{cases} 1, if\; Z_j^t = \arg\min \delta(a_{i,z}), \\ z \epsilon Z^t \\ null.\, otherwise \end{cases}$$

Set $t = t + 1$. If $U^t = U^{t-1}$ or $t = Maximum\ Iteration$ then go to step 4 otherwise go to step 2

4. Calculate the final cluster set $Z^*$ as:

$$Z_{j,l}^* = dl \times \frac{\sum_{a_i \in A_l} u_{i,j}^{t+1} \times A_{i,l}}{\sum_{a_i \in A_l} u_{i,j}^{t+1}} \ \forall l \in \bigcup_{a_i \in C_j^{t+1}} \gamma_{a_i}$$

### 3.7: kNN with AWPD

Use of the AWPD as the root discrepancy, the kNN classifier can be directly applicable to datasets with Non-Existence attributes. There is no need for pre-processes like marginalization or imputation. In this portion, we have discussed about AWPD aided kNN classifier (kNN-AWPD).

Let us find a P $= P_1 \cup P_2$ Dataset, which $P_1 \subset R^m$ and $P_2 \subset R^m$ are the training and testing sets respectively. Let $P_1$ consist of $n_1$ training points $P_{1,i} \in R^m$ (some of have Non-Existence attributes) and let $Q_1$ be the set of correlating class labels $Q_{1,i}$, $\in$ C ($Q_{1,i}$ being the class label of $P_{1,i}$), where C = $\{c1, c2, \cdots, c_l\}$ is the set of all possible class labels. Let $P_2$ contain $n_2$ test instances $P_{2,i} \in R^m$, $i \in \{1, 2, \cdots, n_2\}$.

Furthermore, let $\eta_p^A$ denote the set of k nearest neighbors of a point p among the points in a set A, that is,

$$\eta_p^{A} = \mathop{arg\,max}_{B \subset A, |B| = k} \sum_{y \in B} \delta(P, Q)$$

The class label $Q_{2,i}$ of the test points $P_{2,i}$ is expected as follows after the kNN-AWPD rules of classification:

$$q_{2,i} = \mathop{arg\,min}_{r_{j \in R}} |P_{r,j} \cap \eta_{p_{2,i}}^{p_1}|$$

In other words, kNN-AWPD predicts the class label of a test point $p_{2,i} \in P_2$ to be that of the maximum numbers of points from among the k nearest neighbors of $p_{2,i}$ in the set of training points $P_1$. When multiple class labels occur as many as possible, these relations must be overcome by randomly assigning one of those labels to the test point.

## 4. Results

In this portion, we reported the results of several experiments to validate the validity of the propo-sed K-MEANS++-AWPD and Scalable K-MEANS++-AWPD clustering algorithms and kNN-AWPD classification algorithm. We defined the experimental setup used to test the proposed approaches in the following subsections. The results of the experiments for the K-MEANS++-AWPD algorithm, the Scalable K-MEANS++-AWPD algorithm and kNN-AWPD are respectively presented thereafter. We have presented and discussed the result of four sets of experiment conducted to determine the performance of all these algorithms. The four test sets deal with representations of the respective MCAR, MAR, MNAR-1 and MNAR-2.

### 4.1 Datasets for Clustering

We have taken 10 real-world datasets from the University of California at Irvine (UCI) repository [20], the Jin Genomics Datasets (JGD) repository [21] and Kaggle datasets [22]. Each attribute of each dataset is normalized so as to have zero mean and unit standard deviation. The details of these 10 datasets are listed in Table 1:

**Table 1: Detail of the 10 real datasets for Clustering**

| Dataset | Instances | Attributes | Classes | Repository |
|---------|-----------|------------|---------|------------|
| Iris | 150 | 4 | 3 | KAGGLE |
| Sonar | 208 | 60 | 2 | KAGGLE |
| Glass | 214 | 9 | 6 | KAGGLE |
| Leaf | 340 | 15 | 36 | JGD |
| Seeds | 210 | 7 | 3 | JGD |
| Libras | 360 | 90 | 15 | UCI |
| Chronic Kidney | 800 | 24 | 2 | UCI |

| Vowel Context | 990 | 14 | 11 | UCI |
| Isolate | 1559 | 617 | 26 | UCI |
| Landsat | 6435 | 36 | 6 | UCI |

**4.3 Compare between Direct method and Imputation**

**Table 2: Scalable K-MEANS++ AWPD against MCAR**

| Dataset | Scalable K-MEANS++-AWPD | K-MEANS++-AWPD | K-MEANS FWPD | ZI | MI |
|---|---|---|---|---|---|
| Landsat | **0.100±0.010** | 0.987±0.010 | 0.907±0.020 | 0.789±0.028 | 0.764±0.024 |
| Iris | 0.839±0.100 | **0.849±0.100** | 0.749±0.111 | 0.663±0.118 | 0.116±0.098 |
| Leaf | **0.499±0.025** | 0.489±0.025 | 0.400±0.014 | 0.312±0.021 | 0.321±0.019 |
| Sonar | **0.791±0.105** | 0.777±0.105 | 0.677±0.155 | 0.641±0.198 | 0.625±0.201 |
| Glass | **0.598±0.007** | 0.578±0.007 | 0.478±0.107 | 0.446±0.119 | 0.132±0.072 |
| Seeds | 0.831±0.030 | **0.846±0.030** | 0.806±0.030 | 0.721±0.017 | 0.242±0.039 |
| Libras | **0.796±0.180** | 0.756±0.180 | 0.656±0.080 | 0.631±0.087 | 0.112±0.051 |
| Chronic Kidney | **0.848±0.009** | 0.818±0.009 | 0.798±0.002 | 0.791±0.009 | 0.221±0.013 |
| Vowel Context | **0.489±0.021** | 0.456±0.031 | 0.388±0.031 | 0.352 ± 0.028 | 0.328±0.029 |
| Isolate | 0.609±0.109 | **0.619±0.089** | 0.579 ± 0.117 | 0.523 ± 0.093 | 0.525 ± 0.097 |
| **Best value in the Bold Phase** | | | | | |

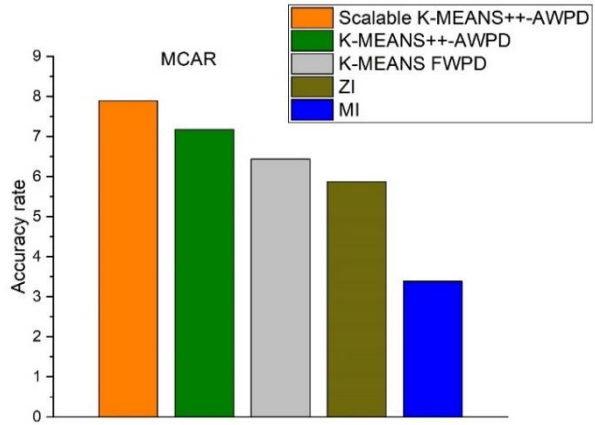

Figure 1: Accuracy rate for Direct and Imputation method against MCAR.

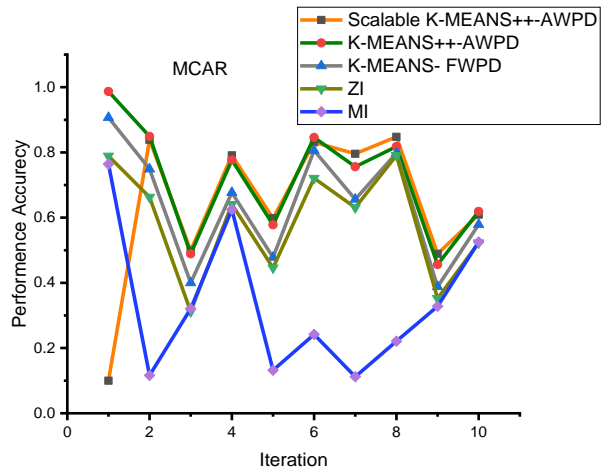

Figure 2: Accuracy rate for all datasets point against MCAR.

The table and graphs represent the accuracy rate and performances accuracy of Non-Existence attributes by MCAR Non-Existence using Scalable K-MEANS++-AWPD, K-MEANS++-AWPD, K-MEANS-FWPD and Imputation method with K-MEANS clustering algorithm. In this phase, we have compared among these four types of methods. We have seen in this graph, Scalable K-MEANS++-AWPD algorithm is mostly probable to K-MEANS-FWPD algorithm, K-MEANS++-

AWPD algorithm and Imputation method. Some of the datasets with K-MEANS++-AWPD has given the best result.

**Table 3: Scalable K-MEANS++-AWPD against MAR**

| Dataset | Scalable K-MEANS++-AWPD | K-MEANS++-AWPD | K-MEANS-FWPD | ZI | MI |
|---|---|---|---|---|---|
| Landsat | **0.954±0.031** | 0.913±0.031 | 0.890±0.050 | 0.769±0.028 | 0.784±0.024 |
| Iris | 0.827±0.096 | 0.837±0.096 | 0.777±0.196 | 0.767±0.186 | **0.841±0.123** |
| Leaf | **0.657±0.055** | 0.557±0.055 | 0.497±0.055 | 0.372±0.121 | 0.421±0.036 |
| Sonar | **0.743±0.211** | 0.690±0.211 | 0.620±0.211 | 0.578±0.067 | 0.569±0.075 |
| Glass | 0.662±0.022 | **0.682±0.022** | 0.592±0.022 | 0.435±0.148 | 0.565±0.122 |
| Seeds | **0.890±0.056** | 0.849±0.056 | 0.779±0.056 | 0.721±0.017 | 0.764±0.022 |
| Libras | **0.741±0.074** | 0.731±0.074 | 0.681±0.074 | 0.661±0.087 | 0.689±0.051 |
| Chronic Kidney | **0.863±0.006** | 0.803±0.006 | 0.793±0.006 | 0.771±0.006 | 0.789±0.051 |
| Vowel Context | **0.507 ± 0.035** | 0.497 ± 0.021 | 0.472 ± 0.044 | 0.461±0.072 | 0.436±0.064 |
| Isolate | **0.754±0.054** | 0.729±0.072 | 0.713±0.051 | 0.691±0.046 | 0.661±0.046 |
| **Best value in the Bold Phase** | | | | | |

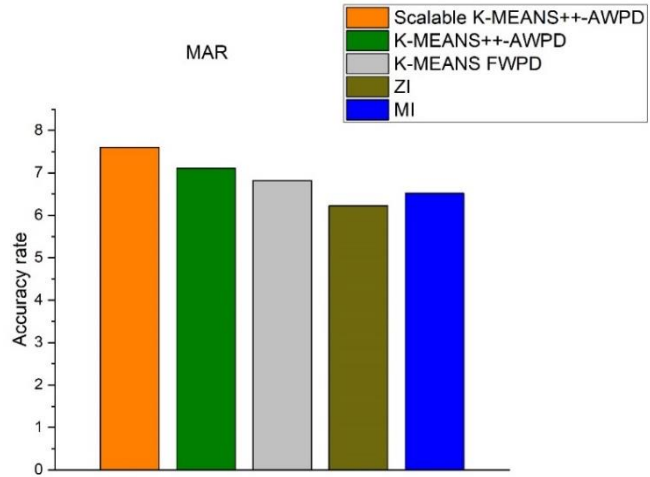

Figure 3: Accuracy rate for Direct and Imputation method against MAR.

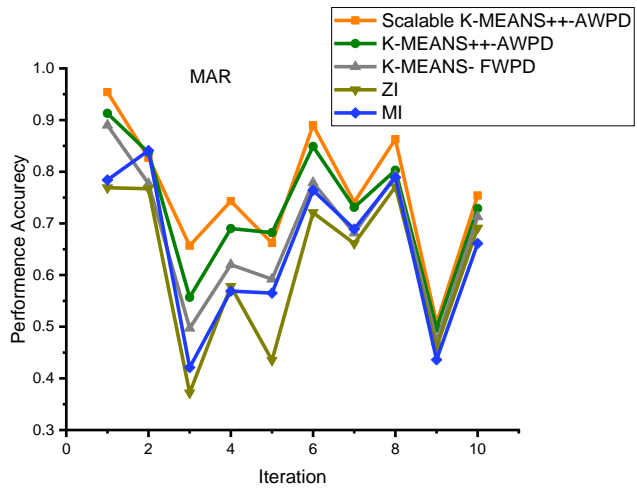

Figure 4: Accuracy rate for all datasets point against MAR.

The table and graphs represent the accuracy rate performances accuracy of Non-Existence attributes by MAR Non-Existence using Scalable K-MEANS++-AWPD, K-MEANS++-AWPD, K-MEANS-FWPD and Imputation method with K-MEANS clustering algorithm. In this phase, we have compared among these four types of methods. We have seen in this graph, Scalable K-MEANS++-AWPD algorithm is mostly probable to K-MEANS-FWPD algorithm, K-MEANS++-AWPD algorithm and Imputation method. Some of the datasets with K-MEANS++-AWPD and Imputation method has given the best result.

**Table 4: Scalable K-MEANS++-AWPD against MNAR-1**

| Dataset | Scalable K-MEANS++-AWPD | K-MEANS++-AWPD | K-MEANS-FWPD | ZI | MI |
|---|---|---|---|---|---|
| Landsat | **0.864±0.014** | 0.787±0.014 | 0.755±0.014 | 0.682±0.128 | 0.703±0.154 |
| Iris | **0.756±0.067** | 0.716±0.067 | 0.681±0.067 | 0.689±0.146 | 0.124±0.066 |
| Leaf | **0.566±0.013** | 0.466±0.013 | 0.425±0.013 | 0.392±0.021 | 0.381±0.026 |
| Sonar | 0.613±0.087 | **0.623±0.087** | 0.599±0.087 | 0.537±0.267 | 0.541±0.275 |
| Glass | 0.453±0.101 | **0.463±0.101** | 0.413±0.101 | 0.381±0.118 | 0.152±0.042 |
| Seeds | **0.802±0.019** | 0.767±0.019 | 0.756±0.019 | 0.781±0.047 | 0.274±0.062 |
| Libras | 0.733±0.121 | **0.743±0.121** | 0.701±0.121 | 0.641±0.071 | 0.352±0.151 |
| Chronic Kidney | 0.757±0.011 | **0.767±0.011** | 0.721±0.011 | 0.704±0.026 | 0.359±0.041 |
| Vowel Context | 0.491±0.051 | 0.473±0.024 | 0.453±0.044 | 0.412±0.049 | 0.435±0.056 |
| Isolate | **0.722±0.051** | 0.708 ± 0.057 | 0.691±0.098 | 0.680±0.103 | 0.680±0.082 |
| **Best value in the Bold Phase** | | | | | |

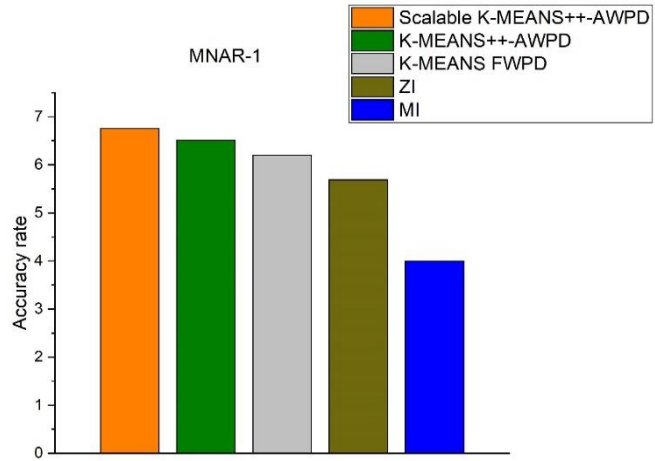

Figure 5: Accuracy rate for Direct and Imputation method against MNAR-1.

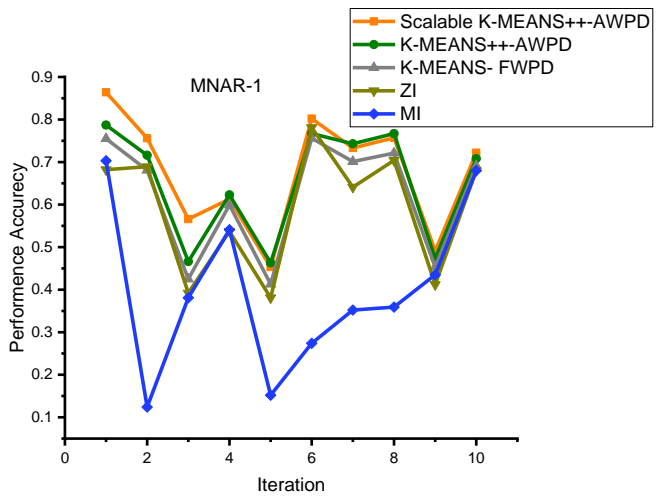

Figure 6: Accuracy rate for all datasets point against MNAR-1.

The table and graphs represent the performance accuracy and accuracy rate of Non-Existence attributes by MNAR-1 Non-Existence using Scalable K-MEANS++-AWPD, K-MEANS++-AWPD, K-MEANS-FWPD and Imputation method with K-MEANS clustering algorithm. In this phase, we have compared among these four types of methods. We have seen in this graph, Scalable K-MEANS++-AWPD algorithm is mostly probable to K-MEANS-FWPD algorithm, K-MEANS++-AWPD algorithm and Imputation method. Some of the datasets with K-MEANS++-AWPD and K-MEANS-FWPD has given the best result.

**Table 5: Scalable K-MEANS++-AWPD against MNAR-2**

| Dataset | Scalable K-MEANS++-AWPD | K-MEANS++-AWPD | K-MEANS-FWPD | ZI | MI |
|---|---|---|---|---|---|
| Landsat | 0.844±0.114 | **0.862±0.114** | 0.821±0.114 | 0.781±0.111 | 0.763±0.074 |
| Iris | **0.831±0.045** | 0.791±0.045 | 0.760±0.045 | 0.707±0.146 | 0.624±0.066 |
| Leaf | 0.461±0.013 | **0.472±0.013** | 0.452±0.013 | 0.385±0.021 | 0.381±0.026 |
| Sonar | **0.786±0.187** | 0.726±0.187 | 0.699±0.187 | 0.637±0.267 | 0.548±0.275 |
| Glass | **0.591±0.121** | 0.561±0.121 | 0.501±0.121 | 0.411±0.118 | 0.407±0.042 |
| Seeds | **0.877±0.019** | 0.857±0.019 | 0.814±0.019 | 0.701±0.047 | 0.734±0.062 |
| Libras | 0.700±0.021 | **0.708±0.021** | 0.698±0.021 | 0.661±0.071 | 0.672±0.151 |
| Chronic Kidney | 0.712±0.115 | **0.720±0.115** | 0.698±0.115 | 0.614±0.026 | 0.721±0.041 |
| Vowel Context | **0.498±0.018** | 0.478±0.048 | 0.458±0.048 | 0.404±0.041 | 0.396±0.064 |
| Isolate | **0.819±0.081** | 0.799±0.061 | 0.789±0.061 | 0.765±0.076 | 0.747±0.056 |
| **Best value in the Bold Phase** | | | | | |

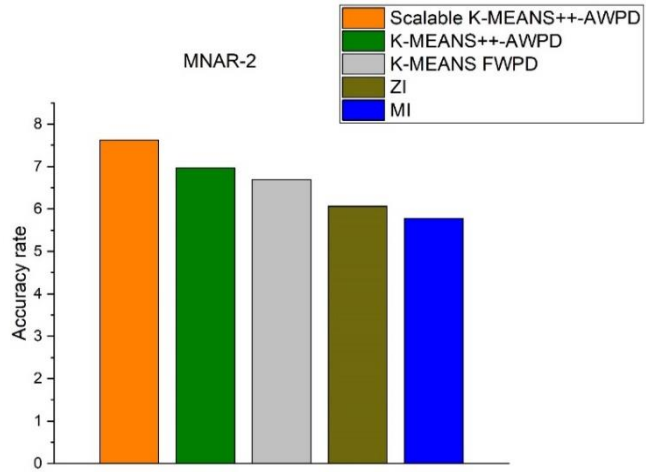

Figure 7: Accuracy rate for Direct and Imputation method against MNAR-2.

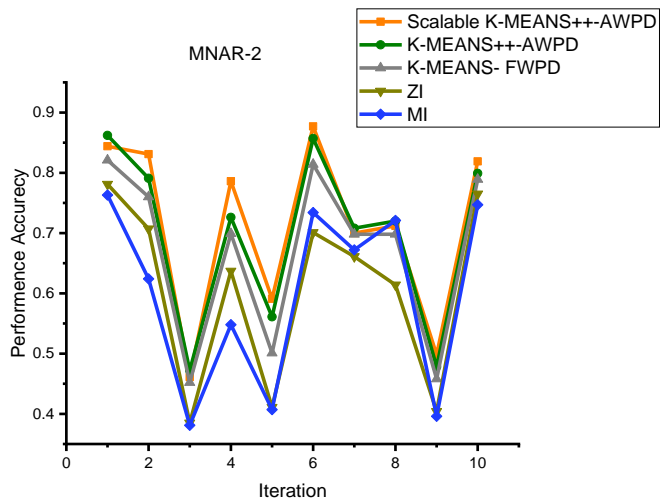

Figure 8: Accuracy rate for all datasets point against MNAR-2.

The table and graphs represent the accuracy rate and performance accuracy of Non-Existence attributes by MNAR-2 Non-Existence using Scalable K-MEANS++-AWPD, K-MEANS++-AWPD, K-MEANS-FWPD and Imputation method with K-MEANS clustering algorithm. In this phase, we have compared among these four types of methods. We have seen in this graph, Scalable K-MEANS++-AWPD algorithm is mostly probable to K-MEANS-FWPD algorithm, K-MEANS++-AWPD algorithm and Imputation method. Some of the datasets with K-MEANS++-AWPD and K-MEANS-FWPD has given the best result.

### 4.4 Datasets and results for classification with PDM or SDM

In this section, we have presented and discussed the results of four sets of experiments conducted to evaluate the performance of the proposed kNN-AWPD method. The four sets of experiments respectively deal with simulations of MCAR, MAR, MNAR-1 and MNAR-2. These are simulated by appropriately removing attributes from 5 datasets, taken from the University of California at Irvine (UCI) repository [23] and KAGGLE datasets [22]. The details of the used datasets are shown in Table 6:

**Table 6: Detail of the 05 real datasets for Classification**

| Dataset | Instances | Attributes | Classes | Repository |
|---------|-----------|------------|---------|------------|
| Glass | 214 | 10 | 6 | KAGGLE |
| Iris | 150 | 4 | 3 | KAGGLE |
| Sonar | 208 | 60 | 2 | UCI |
| Breast Tissue | 106 | 10 | 6 | UCI |
| Bank note | 1372 | 4 | 2 | UCI |

### 4.5 Compare between Direct method (PDM vs SDM) and Imputation method

In data science kNN-AWPD algorithm is the process of directly applied in dataset. The result of the experiments are listed in the term:

**Table7: kNN-AWPD against MCAR for Classification Accuracy**

| Dataset | kNN-AWPD | kNN-FWPD | ZI | MI |
|---|---|---|---|---|
| Glass | **0.753±0.043** | 0.654±0.087 | 0.551±0.060 | 0.532±0.069 |
| Sonar | **0.784±0.026** | 0.698±0.018 | 0.597±0.054 | 0.569±0.087 |
| Iris | **0.776±0.013** | 0.723±0.003 | 0.671±0.009 | 0.635±0.098 |
| Breast Tissue | **0.629±0.251** | 0.529±0.321 | 0.551±0.434 | 0.592±0.256 |
| Bank note | **0.554±0.065** | 0.439±0.007 | 0.439±0.007 | 0.439±0.007 |
| **Best value in the Bold Phase** | | | | |

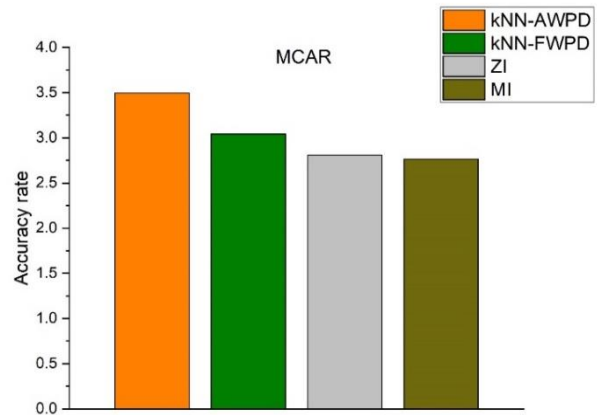

Figure 9: kNN-AWPD against MCAR for Classification Accuracy.

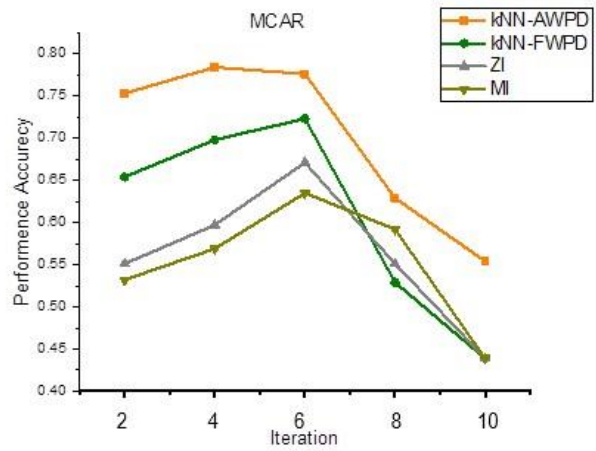

Figure 10: kNN-AWPD against MCAR for Classification Accuracy.

The table and graphs represent the performance accuracy and accuracy rate of Non-Existence attributes by MCAR Non-Existence using kNN-AWPD and kNN-FWPD, Imputation method with kNN classification algorithm. In this phase, we have compared among all types of algorithms. We have seen in this graph, kNN-AWPD is mostly probable to all. But some of the dataset are probable by kNN-FWPD measure.

**Table 8: kNN-AWPD against MAR for Classification Accuracy**

| Dataset | kNN-AWPD | kNN-FWPD | ZI | MI |
|---|---|---|---|---|
| Glass | **0.705±0.104** | 0.676±0.121 | 0.649±0.054 | 0.658±0.019 |
| Sonar | **0.712±0.055** | 0.631±0.025 | 0.597±0.054 | 0.597±0.054 |
| Iris | **0.800±0.071** | 0.732±0.011 | 0.671±0.009 | 0.635±0.098 |
| Breast Tissue | **0.786±0.009** | 0.699±0.009 | 0.622±0.034 | 0.687±0.106 |
| Bank note | **0.743±0.044** | 0.651±0.068 | 0.548±0.010 | 0.548±0.010 |
| **Best value in the Bold Phase** | | | | |

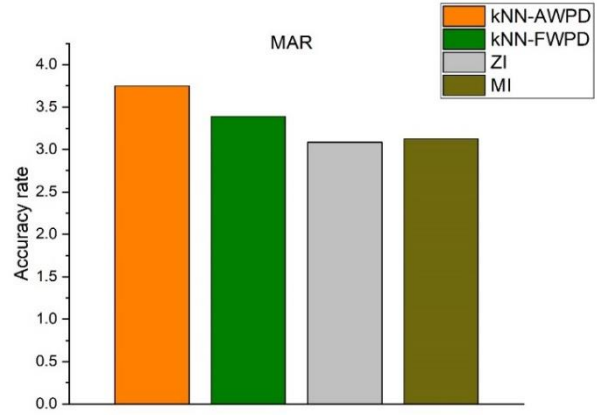

Figure 11: kNN-AWPD against MAR for Classification Accuracy.

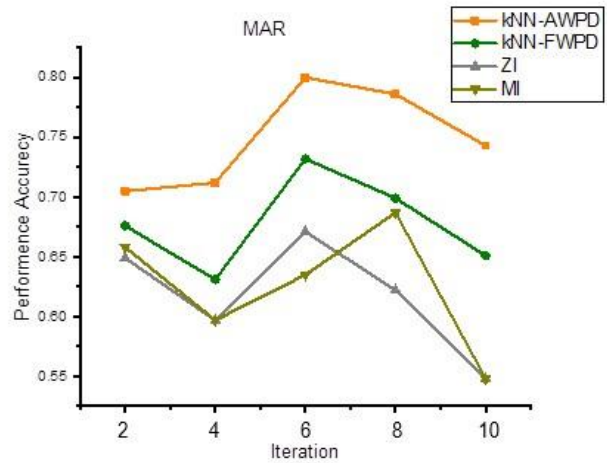

Figure 12: kNN-AWPD against MAR for Classification Accuracy.

The table and graphs represent the accuracy rate and performance accuracy of Non-Existence attributes by MAR Non-Existence using kNN-AWPD and kNN-FWPD, Imputation method with kNN classification algorithm. In this phase, we have compared among all types of algorithms. We have seen in this graph, kNN-AWPD is mostly probable to all. Some of the dataset are given best result in kNN-FWPD measure.

**Table 9: kNN-AWPD against MNAR-1 for Classification Accuracy**

| Dataset | kNN-AWPD | kNN-FWPD | ZI | MI |
|---------|----------|----------|-----|-----|
| Glass | **0.676±0.111** | 0.601±0.102 | 0.573±0.153 | 0.601±0.021 |
| Sonar | **0.686±0.108** | 0.598±0.121 | 0.537±0.104 | 0.551±0.008 |
| Iris | **0.801±0.008** | 0.709±0.013 | 0.691±0.013 | 0.691±0.013 |
| Breast Tissue | **0.843±0.079** | 0.721±0.009 | 0.682±0.034 | 0.612±0.106 |
| Bank note | **0.787±0.036** | 0.695±0.021 | 0.614±0.121 | 0.561±0.110 |
| **Best value in the Bold Phase** | | | | |

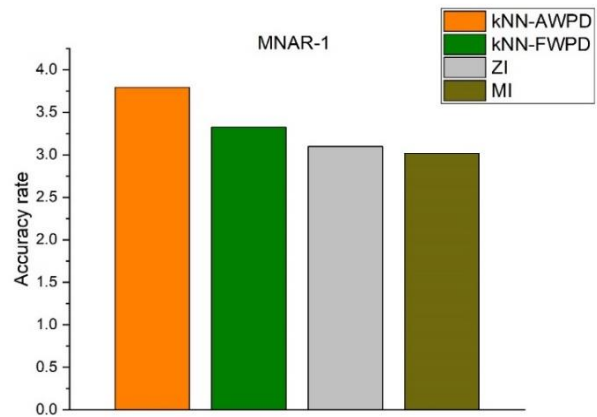

Figure 13: kNN-AWPD against MNAR-1 for Classification Accuracy.

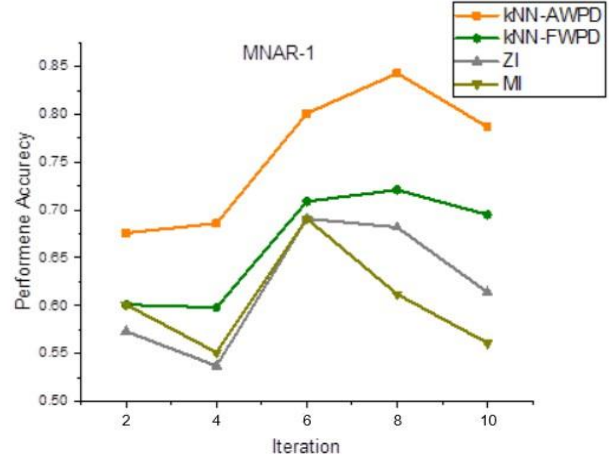

Figure 14: kNN-AWPD against MNAR-1 for Classification Accuracy.

The table and graphs represent the accuracy rate and performance accuracy of Non-Existence attributes by MNAR-1 Non-Existence using kNN-AWPD and kNN-FWPD, Imputation method with kNN classification algorithm. In this phase, we have compared among all types of algorithms. We have seen in this graph, kNN-AWPD is fully probable to all.

**Table 10: kNN-AWPD against MNAR-2 for Classification Accuracy**

| Dataset | kNN-AWPD | kNN-FWPD | ZI | MI |
|---|---|---|---|---|
| Glass | **0.857±0.044** | 0.769±0.020 | 0.712±0.004 | 0.684±0.125 |
| Sonar | **0.739±0.065** | 0.685±0.029 | 0.657±0.031 | 0.597±0.054 |
| Iris | **0.708±0.010** | **0.708±0.010** | 0.689±0.004 | 0.645±0.087 |
| Breast Tissue | **0.715±0.010** | 0.664±0.011 | 0.701±0.011 | 0.687±0.007 |
| Bank note | **0.611±0.021** | **0.611±0.021** | 0.548±0.010 | 0.578±0.073 |
| **Best value in the Bold Phase** | | | | |

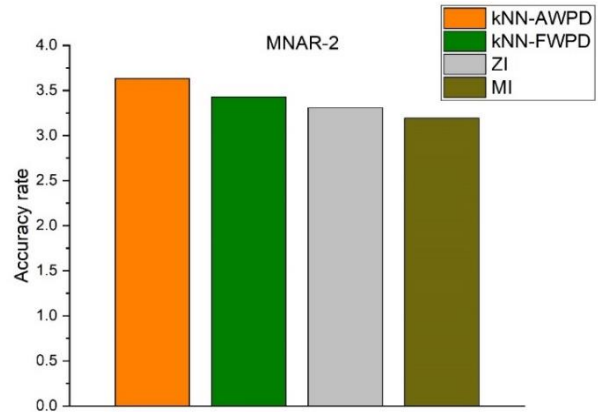

Figure 15: kNN-AWPD against MNAR-2 for Classification Accuracy.

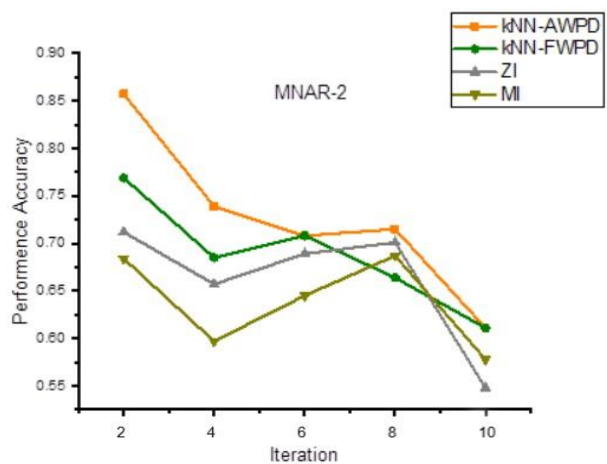

Figure 16: kNN-AWPD against MNAR-2 for Classification Accuracy.

The table and graphs represent the performance accuracy and accuracy rate of Non-Existence attributes by MNAR-2 Non-Existence using kNN-AWPD and kNN-FWPD, Imputation method with kNN classification algorithm. In this phase, we have compared among all types of algorithms. We have seen this graph, kNN-AWPD is fully probable to all. But some of the datasets has given same result with kNN-FWPD.

## 5. Conclusion

In this research, we have proposed to use the AWPD measure as a viable alternative to imputation and marginalization approaches to handle the problem of non-existence attributes in data clustering and classification. The proposed measure attempts to estimate the original duration of each other data points by adding a penalty term to those pair-wise distances which cannot be calculated on the entire attribute space due to non-existence attributes. Therefore, unlike existing methods for handling non-existence attributes, AWPD is also able to distinguish between distinct data points which look identical due to Non-Existence attributes. Yet, AWPD also ensures that the Discrepancy for any data instance from itself is never greater than its Discrepancy from any other point in the dataset. Intuitively, these advantages of AWPD should help us better model the original data space which may help in achieving better clustering performance on the incomplete data. Therefore, we have used the proposed our AWPD measure to put forth the K-MEANS++-AWPD and scalable K-MEANS-AWPD is clustering algorithm and kNN-AWPD is classification algorithm, which are applicable explicitly for datasets with non-existence attributes. We have conducted extensive experimentation on the new techniques using various benchmark datasets and found the new approach to produce generally better results for partition compared with few of them the general imputation approaches which are generally used to control of the non-existence attributes problem. In fact, it is observed from the experiments that the implementation of the schemes of imputation varies with category of non-existence and the algorithm used for clustering and classification. The proposed approaches, in the other side, exhibits good performance across all types of Non-Existence as well as partition clustering paradigms. The experimental results attest to the ability of AWPD to better model the original data space, compared to existing methods.

However, it must be estimated that, the performance of all these methods, including the AWPD based ones, can vary depending on the structure of the dataset concerned, the choice of the proximity measure used, and the pattern and size of Non-Existence plaguing the data. Fortunately, $\beta$ parameter embedded in AWPD can be varied in accordance with the extent of Non-Existence to achieve desired results. The results section indicates that it may be useful to choose a high value of $\beta$ when a massive fraction of the attributes are unobserved, and to choose a smaller value when only a few of the attributes are Non-Existence . However, in the presence of a sizable amount of Non-Existence and the absence of ground-truths to validate the merit of the achieved clustering, it is safest to choose a value of β proportional to the percentage of Non-Existence attributes restricted within the range [0.1, 0.25] [1].

We will present an appendix dealing with an extension of the AWPD measure to problems with absent attributes and show that this modified form of AWPD is a semi-metric (Structural Non-Existence). After that, we will minimize time complexity of this research work.